\documentclass[10pt,twocolumn,letterpaper]{article}

\usepackage[pagenumbers]{cvpr} 

\usepackage{amsmath}
\usepackage[most]{tcolorbox} 
\usepackage{booktabs}
\usepackage{tabularx}
\usepackage{multirow}

\usepackage{array}
\usepackage{amssymb}
\usepackage{subcaption}
\usepackage{wrapfig}
\usepackage{tikz} 

\definecolor{cvprblue}{rgb}{0.21,0.49,0.74}
\usepackage[pagebackref,breaklinks,colorlinks,allcolors=cvprblue]{hyperref}

\newcommand{\YSR}[1]{\textcolor{blue}{}}
\newcommand{\LX}[1]{\textcolor{orange}{}}

\def\paperID{16641} 
\def\confName{CVPR}
\def\confYear{2025}

\title{DIFFER: Disentangling Identity Features via Semantic Cues \\for Clothes-Changing Person Re-ID}

\author{  
Xin Liang \quad Yogesh S Rawat \\
Center for Research in Computer Vision, University of Central Florida \\
{\tt\small xin.liang@ucf.edu}, {\tt\small yogesh@ucf.edu} \\
}

\begin{document}
\maketitle
\begin{abstract}
Clothes-changing person re-identification (CC-ReID) aims to recognize individuals under different clothing scenarios. Current CC-ReID approaches either concentrate on modeling body shape using additional modalities including silhouette, pose, and body mesh, potentially causing the model to overlook other critical biometric traits such as gender, age, and style, or they incorporate supervision through additional labels that the model tries to disregard or emphasize, such as clothing or personal attributes. However, these annotations are discrete in nature and do not capture comprehensive descriptions.

In this work, we propose DIFFER: Disentangle Identity Features From Entangled Representations, a novel adversarial learning method that leverages textual descriptions to disentangle identity features. Recognizing that image features inherently mix inseparable information, DIFFER introduces NBDetach, a mechanism designed for feature disentanglement by leveraging the separable nature of text descriptions as supervision. It partitions the feature space into distinct subspaces and, through gradient reversal layers, effectively separates identity-related features from non-biometric features. We evaluate DIFFER on 4 different benchmark datasets (LTCC, PRCC, CelebreID-Light, and CCVID) to demonstrate its effectiveness and provide state-of-the-art performance across all the benchmarks. DIFFER consistently outperforms the baseline method, with improvements in top-1 accuracy of 3.6\% on LTCC, 3.4\% on PRCC, 2.5\% on CelebReID-Light, and 1\% on CCVID. Our code can be found \href{https://github.com/xliangp/DIFFER.git}{here}.
\LX{[done]}\YSR{one complain was that the comparison is not fair with sota, maybe here we can say DIFFER improves XYZ over the baseline with the help of these semantic pseudo labels... and also provide sota performance on all benchmarks... something like this, DIFFER consistently improves the ReID performance over baseline with the help of semantic pseudo labels and provide state-of-the-art performance across all 4 benchmarks... we never say we are comparing with sota ... think about it...}

\end{abstract}    
\section{Introduction}
\begin{figure}[t]
    \centering
    \includegraphics[width=0.8\linewidth]{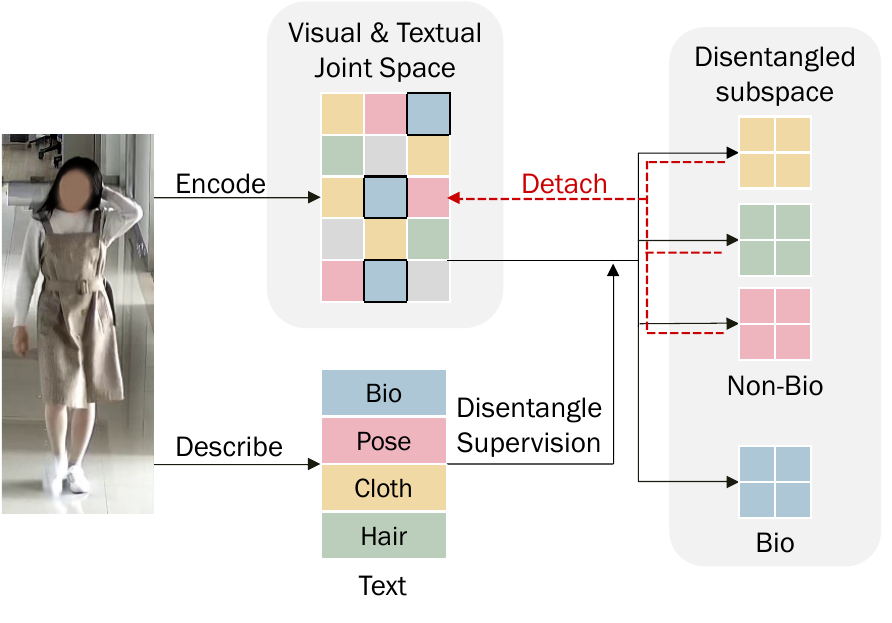}
    \caption{\textbf{\textit{Motivation behind DIFFER}}. Encoded image features often entangle a multitude of information. By leveraging the separable nature of text descriptions, we guide the model to disentangle the feature space into distinct subspaces. This allows us to detach non-biometric factors and preserve the crucial biometric features.
    \LX{[done]}\YSR{ this is a nice figure... can you increase the font size, so that the text is readable? can we do something similar in the main architecture too? nbdetach can be more explained there... its very simplistic at this time... think about this...}
    \LX{[done]}\YSR{ hide/blur the face...}
    \YSR{please share the source for this and other figures.. will see if get time...: }\LX{[from LTCC dataset. The other from CUHK. Should we comment in somewhere?]}}
    \label{fig:motivation}
    \vspace{-0.8cm}
\end{figure}

Person re-identification (ReID) \cite{li_clip-reid_2022,he_instruct-reid_2023} is an important task in computer vision, aiming to match individuals across different camera views and environments. ReID plays a significant role in various real-world applications, including surveillance, security, and intelligent transportation systems. Traditional ReID methods primarily rely on visual cues such as appearance and clothing, assuming that individuals wear the same outfit across different observations. However, this assumption is often unrealistic, particularly in real-world settings where people frequently change their clothes. Clothes-changing re-identification (CC-ReID) studies a more challenging problem for real-world situations, where the goal is to recognize individuals despite changes in their clothing. In this work, we focus on CC-ReID to identify the person over a long time period and in a more dynamic environment.

Existing works on CC-ReID have introduced various strategies to tackle the challenges posed by significant changes in appearance. Many multi-modal approaches rely on external modalities, such as pose \cite{qian2020long}, silhouette \cite{hong2021fine}, clothing removal \cite{mu2022learning}, or body shape modeling \cite{liu2023learning}, to capture identity-related features. While these methods can improve performance, they introduce additional computational overhead during inference and are often prone to errors due to the dependency on external models, which may fail in complex or unconstrained environments. Moreover, by largely relying on body shape information, these models may not adequately capture other pertinent biometric information such as race, gender, and clothing style. 

To overcome these challenges, some recent works rely only on RGB images and leverage clothing labels \cite{gu2022clothes} or personal attributes \cite{lee2022attribute,liu2024distilling,peng2024masked} to learn biometric features. Such discrete annotations are useful for learning biometric features, but
certain non-biometric factors, such as clothing, hairstyles, or postures, which exhibit infinite variability, are not amenable to precise categorical labeling. 
Furthermore, relying only on the elimination of clothes information from the encoded feature could let the model neglect other distracting factors such as hairstyle, pose, or environment.
In this work, we aim to address some of these limitations by proposing a method that disentangles biometric and non-biometric features using descriptive pseudo-labels without relying on additional external modalities during inference, enhancing flexibility.

We propose DIFFER, a framework leveraging multi-modal foundation models to disentangle non-biometric features from identity-specific representations. DIFFER builds on the capabilities of visual-language models (VLMs) to align visual and language semantics, effectively separating biometric and non-biometric features through separable textual descriptions generated from a VLM. Our proposed NBDetach module uses gradient reversal to remove non-biometric information during training, preserving only identity-specific features. Importantly, semantic descriptions are only needed during training and are not required at inference, enhancing practicality. Evaluated on four benchmark datasets for clothes-changing person re-identification (CC-ReID), DIFFER consistently outperforms baseline methods without needing additional modalities, maintaining both flexibility and robustness.

The contributions of this work are as follows: 
(1) We propose DIFFER, a novel multimodal approach for disentangling non-biometric information while preserving biometric features via the guidance of separated semantic descriptions. \LX{done}\YSR{we want to highlight multimodal part? }
(2) We introduce the use of semantic descriptions as pseudo-labels for biometric and non-biometric attributes, generated using VLMs. 
(3) We develop NBDetach, a gradient reversal-based technique for feature disentanglement that enables the model to ignore non-biometric features. 
(4) We conduct extensive evaluations across multiple datasets to validate the robustness and scalability of DIFFER.


\section{Related Works}
\noindent
\textbf{Clothes Changing Person Re-Identification.}
CC-ReID proposes a unique problem to identify the same person in the long term,  various outfits, and dynamic environments. An increasing amount of research contributes to this problem,  which can be roughly categorized into single-modality and multi-modality methods.

Methods using only RGB modality usually introduce additional supervision signals to extract the cloth-irrelevant features, such as clothing labels  \cite{gu2022clothes,yang2023good,chan2023learning,han2023clothing}, person attributes  \cite{peng2024masked} and descriptions  \cite{liu2024distilling}. Multi-modality methods usually introduce human parsing  \cite{mu2022learning,cui2023dcr,guo2023semantic}, 3D shapes \cite{chen2021learning,liu2023learning}, or silhouette \cite{hong2021fine,nguyen2024contrastiveClothing} to enforce the model to ignore the cloth-irrelevant information and focus on the body shape. These methods transform the problem from eliminating interference factors into cross-modality feature alignment, but these additional modalities may eliminate crucial visual information such as age, gender, or style.

\begin{figure*}[t!] 
  \centering
  \includegraphics[width=0.9\textwidth]{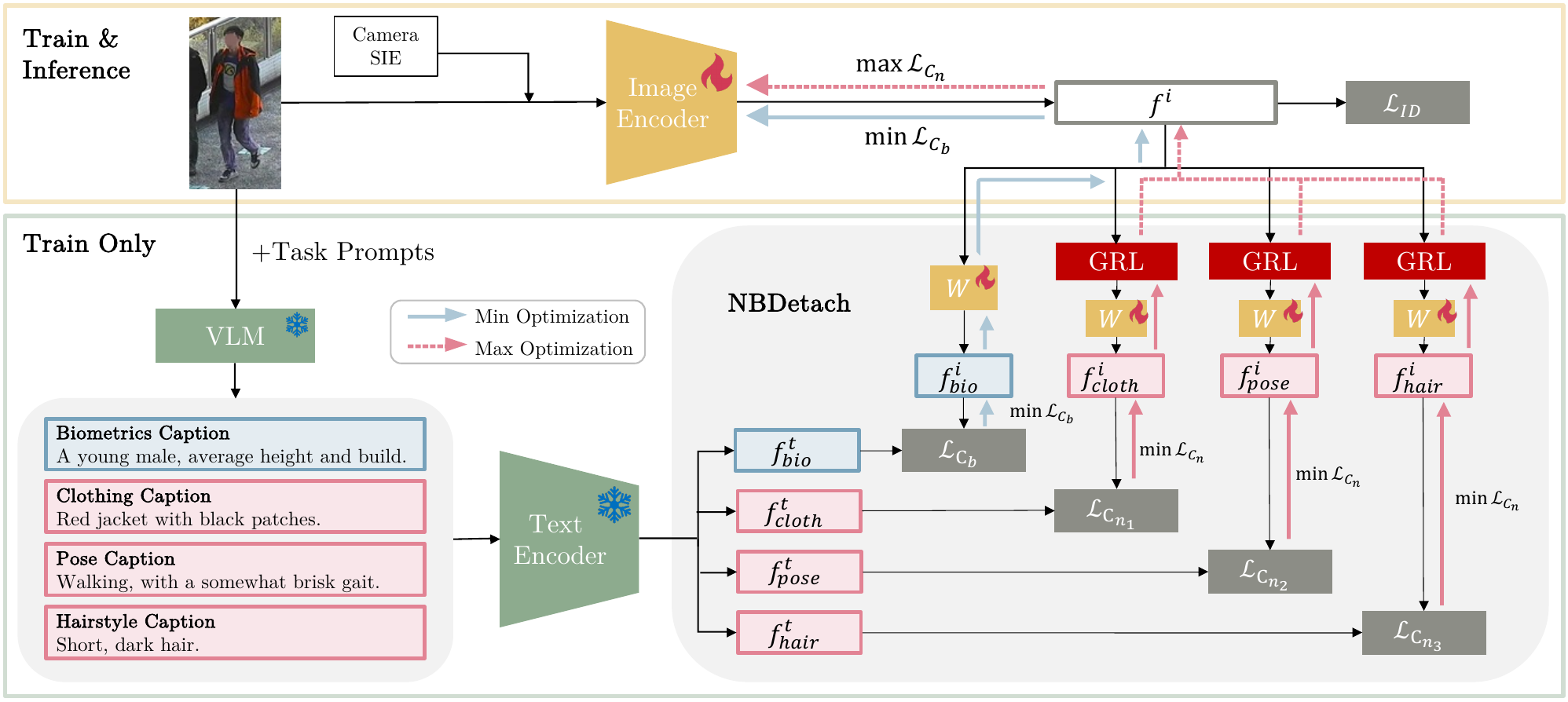} 
  \caption{\textbf{\textit{Overview of DIFFER:}} First, we use VLM with DIFFERent task prompts to generate DIFFERent textual captions in the perspective of biometrics and non-biometrics factors, which are encoded as textual features, $f^t_{b}, f^t_{n_1},...,f^t_{n_N}$. Next, the image is encoded as image feature $f^i$ in an entangled feature space, and camera side information embedding (Camera SIE) is added during positional embedding. Finally, \textbf{NBDetach} is proposed to detach the non-biometric information and preserve the biometric information. Fully linear projection layers $W$ are used to project the image feature to DIFFERent subspaces. Then \textbf{GLR} (Gradient Reverse Layer) is added to disentangle the non-biometric information from the entire feature space supervised by non-biometric contrastive losses $\mathcal{L}_{\text{C}_{n}}$.  
  \YSR{in vlm + task prompt, keep vlm inside, task prompt should be input to vlm...}
  \YSR{for image encoder, can we just show a traditional transformer model at high level, showing image patches, tokesn etc... just to fill the white space... and we can also show what is the output, patch tokens or cls token etc... only at a high level... not too detailed.. but little more details then this... it will be good if we can show feautures like you have shown in figure 1, and then use them for disentanglement... and add more details on grl... lot of white space to use...}
  \YSR{use DIFFERent shape/style types for DIFFERent aspects, you have model, features, trainable weights, loss, etc...}
  }
  \label{fig:method}
\end{figure*}

\noindent
\textbf{Foundation Models.}
 Foundation Models are large-scale models that are pre-trained on vast amounts of diverse data and can be adapted to a wide range of downstream tasks through fine-tuning or prompt-based techniques. Large language models (LLMs), such as GPT4 \cite{achiam2023gpt}, have demonstrated impressive generalization across a wide range of natural language processing tasks. Vision-language models (VLM) are pre-trained to match image and language on large scales of data, allowing them to perform tasks involving both texts and images in a joint feature space, such as LLaVA \cite{liu2023llava},  InternVL \cite{chen2024expanding} and CogVLM \cite{wang2023cogvlm}. CLIP (Contrastive Language-Image Pretraining) models \cite{radford2021learning} employ contrastive learning to learn joint representations of images and texts. For example, EVA-CLIP-18B \cite{sun_eva-clip-18b_2024} successfully scales up to 18 billion parameters and is the largest and most powerful open-source CLIP model to date. 
 
 Extensive research has successfully applied the pre-trained VLMs to fine-grained image re-identification tasks, including \cite{li_clip-reid_2022,he_instruct-reid_2023,yang_mllmreid_2024}. However, some existing methods failed to utilize the description ability of VLM from DIFFERent perspectives and focus only on one aspect of the VLM's outputs, such as body shape \cite{liu2024distilling}; other methods\cite{li_clip-reid_2022} require prompt learning and a two-stage training process, limiting their efficiency and flexibility. \LX{done}\YSR{rephrase this, also clip-reid is not person description, you are doing person description, clip-red is learning prompts...}

\noindent
\textbf{Feature disentanglement.}
Feature disentanglement \cite{wang_disentangled_2024-1} aims to isolate and separate distinct factors within the embedded representation space. Traditional methods often rely on generative models like VAEs and GANs \cite{liu_multi-task_2018} to split the latent space into various generative factors, aiding in structured factor separation. Recent advancements have extended feature disentanglement beyond generation tasks to areas such as image classification \cite{feng2019self,sanchez2020learning}, natural language processing \cite{cheng2020improving,zhang_disentangling_2021}, and multimodal applications \cite{materzynska_disentangling_2022}. Techniques vary widely, including multi-task adversarial networks that use min-max optimization for simultaneous encoder and style discriminator training \cite{liu_multi-task_2018}, metric learning-based approaches like DFR \cite{cheng2023disentangled} utilizing gradient reversal layers \cite{ganin2015unsupervised}, and orthogonal linear projections for separating visual and textual features in CLIP \cite{materzynska_disentangling_2022}.

In person re-identification (Re-ID), feature disentanglement has been effectively utilized to separate identity from non-identity factors \cite{bai_disentangled_2020,chan2023learning,azad2024activity,li2024disentangling}. Some Re-ID methods integrate a generative task to separately learn identity and non-identity features \cite{chan2023learning}, while others in domains like vehicle Re-ID use networks like DFLNet \cite{bai_disentangled_2020} to learn orientation-specific and general features simultaneously. For occluded person Re-ID, ProFD \cite{cui2024profd} uses text prompts to disentangle body part features. In clothes-changing Re-ID, dual-stream frameworks have been adopted to isolate identity features from appearance changes and camera variations \cite{li2024disentangling}. However, such methods often require generative models \cite{chan2023learning} or human annotations \cite{bai_disentangled_2020}, highlighting the need for more flexible and annotation-free disentanglement techniques in Re-ID. So we propose DIFFER which uses text descriptions from VLM to guide the model  in disentangling the biometric feature from the visual feature space.




\begin{figure*}[t!]
    \centering
    \small
\begin{tcolorbox}[
  enhanced,
  attach boxed title to top left={yshift=-2mm, xshift=2mm},
  colback=gray!4,
  colframe=black!75!white,
  colbacktitle=black!75!white,
  title=Example Conversation with VLM,
  fonttitle=\bfseries,
  boxed title style={size=small,colframe=black!75!white} ]
  \begin{center}
   \textbf{Biometric Description}
  \end{center} 
  \textbf{Input}: \textit{Describe the individual's overall \textbf{physical appearance}, including estimated age, gender, height (e.g., short, average, tall based on surroundings if applicable), and build(e.g., slender, average, robust).}
  
    \textbf{VLM}:\textit{ The individual appears to be a young male, possibly in his late teens to early twenties. He has short, dark hair and glasses. Based on the surroundings, he seems to be of average height and build.}
    \begin{center}
        \textbf{Non-Biometric Description}
    \end{center}

    \textbf{Input}: \textit{Detail the type of \textbf{clothing} the person is wearing(the style, colors, and any visible logos or patterns), shoes and any accessories (e.g., glasses, watches, jewelry).}
    
    \textbf{VLM}: \textit{The individual is wearing a red jacket with black and white patches. Underneath the jacket, he has a blue shirt with a white logo or emblem on it. He is also wearing purple pants and black and white shoes. He is wearing glasses. 
    }    
\end{tcolorbox}
\end{figure*}

\section{Methodology}

\subsection{Overview}

Given an input image, our goal is to learn a representation that can reliably identify an individual across variations in clothing, poses, environments, and hairstyles. To achieve this, we propose DIFFER: Disentangle Identity Features From Entangled Representations via separated semantic descriptions. DIFFER isolates identity-related (biometric) features from non-identity (non-biometric) factors, as shown in \cref{fig:method}. It utilizes a multi-modal approach where semantic information from textual descriptions is used to disentangle biometric and non-biometric information. DIFFER’s architecture comprises three main components: (1) a visual encoder, (2) semantic learning, and (3) identity feature disentanglement.

A person's image $\mathbf{x}$ is first processed by the visual encoder, producing image features $\mathbf{f}^i$ used for identification. To extract semantic features, we generate biometric and non-biometric descriptions from $\mathbf{x}$ with corresponding prompts from a pre-trained visual-language model (VLM). The resulting textual descriptions are then encoded by a text encoder $\mathcal{T}$, producing textual features $\mathbf{f}^t_{b}$ for biometrics and $\mathbf{f}^{t}_{n_1},...,\mathbf{f}^{t}_{n_N}$ for $N$ distinct non-biometric factors. 

To further disentangle non-biometric factors, we introduce the non-biometric detach module (NBDetach), which detaches non-biometric components while preserving biometric information. NBDetach projects image features $\mathbf{f}^i$ into separate visual biometric features $\mathbf{f}^{i}_{b}$ and visual non-biometric features $\mathbf{f}^{i}_{n_1},...,\mathbf{f}^{i}_{n_N}$, aligning them with corresponding semantic features using contrastive losses $\mathcal{L}_C$.  The gradient reversal layers are added before the non-biometric projection heads, serving as a forgetting mechanism that helps the image encoder remove non-biometric information. During inference, only the visual encoder is needed to obtain biometric features, while the text encoder and NBDetach module are not required.

\YSR{did some changes in previous 3 paras, see if all is good... symbols etc...}


\subsection{Visual encoder}
The visual encoder $\mathcal{I}$ is responsible for extracting visual features from the input image. It encodes the image to generate an entangled representation that contains a mixture of information, including biometric, non-biometric and environmental information. This feature representation serves as the foundation for further disentanglement and identification. To facilitate disentanglement using semantic descriptions, it is important that the latent feature space of visual encoder aligns with the text encoder. We use ViT, a transformers-based visual encoder from a multimodal model that has been pre-trained on image-language pairs \cite{sun_eva-clip-18b_2024}. The classification token of ViT is used to represent image features $\mathbf{f}^i$. 

\subsection{Semantic learning}
Existing clothing-changing ReID methods usually adopt a supervised method to guide the model in learning the desired biometric information, such as the 3D body shape, silhouettes, or personal attributes, while forgetting unnecessary non-biometric information.  However, some non-biometrics labels are hard to acquire since we cannot categorize the infinite possibilities into finite categories, such information including varying clothing styles, poses, or hairstyles. Besides, it is hard to differentiate the image feature encoded by a neural network from different factors in an embedded space. These difficulties have motivated us to utilize the strength of a large visual-language foundation model, which has the ability to describe a person's appearance in an image, and it is easy to separate the text descriptions into different aspects.

In this study, we adopt a context-objective framework to generate our task prompts for the biometric and non-biometric information with a large language model GPT4 \cite{achiam2023gpt}. Then we use an open VLM model, CogVLM \cite{wang2023cogvlm}, to extract the text captions for each image in the training dataset. Our biometric information includes body build, age, gender, etc. This is included in one sentence and is later encoded by a text encoder $\mathcal{T}$ as a biometrics text feature $\mathbf{f}^t_b$. Moreover, the non-biometric descriptions can include hairstyle, clothing, pose, and environment captions with specially designed prompts to extract different information from the image (examples in supplementary material).  The non-biometric textual feature is encoded as  $\mathbf{f}^{t}_{n}$ by the text encoder $\mathcal{T}$. Our task prompts and example descriptions for biometrics and non-biometrics corresponding to the image in \cref{fig:method} are shown in the text box above.

After generating all the text descriptions, we use a summarization prompt to combine all the biometric descriptions for one individual into one summarising text. So all the images belonging to the same individual should have the same biometric text feature. GPT4 is used to summarize the textual descriptions. 
\YSR{do we have any ablation where we show that this summarization is helpful? any experimental results without this summarization?}\LX{the experiment results are not good, so I did not emphasize this part. Could include summarized ablation in suplementary}




\subsection{NBDetach: Feature disentanglement}
To detach the non-biometric information from the image feature space while preserving biometric information, we propose a Non-Biometric Detach (\textbf{NBDetach}) module. The aim is to preserve the biometric features and ignore the non-biometric features. It relies on semantic features from the textual descriptions to guide this disentanglement. 
After the image $\mathbf{x}$ is encoded as the image feature $\mathbf{f}^{i}$ by the image encoder $\mathcal{I}$, we project the image feature into different subspaces to match the biometric and $N$ different non-biometric textual features. The resulting image features are denoted as $\mathbf{f}^{i}_{b}$ for biometric feature and $\mathbf{f}^{i}_{n_k}$ for non-biometric features, where $k\in \{1,...,N\}$. The biometric projection head is denoted as $H_b$, and the non-biometric projection heads are denoted as $H_{n_k}$, where linear projection layers $W$ are used to perform the projections. 
We use an image-to-text contrastive loss \cite{radford2021learning} between textual and visual features for both biometric and non-biometric features.
\begin{equation}
\mathcal{L}_{C_h} = -\frac{1}{B} \sum_{k=1}^{B} \log \frac{\exp(\text{cos}(\mathbf{f}^{i}_{h,k}, \mathbf{f}^{t}_{h,k}) )}{\sum_{j=1}^{B} \exp(\text{cos}(\mathbf{f}^{i}_{h,j}, \mathbf{f}^{t}_{h,j}) )},
\end{equation}
where $h$ is either the biometric $\mathcal{L}_{\text{C}_{b}}$ or non-biometric $\mathcal{L}_{\text{C}_{n}}$, $B$ is the batch size, $\mathbf{f}^{i}_{h,k}$ is the projected image feature for instance $k$, and $\mathbf{f}^{t}_{h,k}$ is the corresponding text feature in domain $h$. 
Our goal is to preserve biometric information in the image feature  $\mathbf{f}^{i}$ and forget non-biometric information. A contrastive loss will help in preserving biometric information in the image features $\mathbf{f}^{i}$, but we want to disentangle the non-biometric information. Therefore we rely on gradient reversal for non-biometric features, which will help detach the non-biometric information from image feature  $\mathbf{f}^{i}$. 
The training objectives are formulated as,

\begin{equation}
\min\limits_{\mathcal{I},H_{b}}\mathcal{L}_{\text{C}_{b}}
\end{equation}
\begin{equation}
\max\limits_{\mathcal{I}} \min\limits_{H_{n}}\mathcal{L}_{\text{C}_{n}}. 
\end{equation}

As shown above, $\mathcal{I}$ and  $H_b$ are optimized jointly to minimize the biometric contrastive loss such that the image feature preserves more biometric information. Conversely, $\mathcal{I}$ and $H_{n}$ play an adversarial role such that the non-biometric projection tries to extract the corresponding non-biometric information as much as possible, while the image encoder tries to detach the non-biometric information from the encoded image feature space. 

To perform the min-max optimizations simultaneously, a gradient reverse layer (GRL) \cite{ganin2015unsupervised} is added between the image encoder $\mathcal{I}$ and the non-biometric projection heads $H_{n}$. GRL multiplies the gradient with a negative coefficient to reverse the gradient direction, ensuring that the training objectives before and after applying the layer are opposite. So it is possible to jointly train the encoder and non-biometric heads with opposite objectives.

\subsection{Loss Function}
\textbf{ID Loss} For the identification loss, we used a standard cross entropy classification loss $\mathcal{L}_{cls}$ and triplet loss $\mathcal{L}_{tri}$. For image feature  $\mathbf{f}^{i}_i$ and its corresponding identity label $y_i$, the ID loss is calculated as:
\begin{gather}
    \mathcal{L}_{cls} =  - \frac{1}{B} \sum_{i=1}^{B} \sum_{c=1}^{C} y_{i,c} \log(\hat{y}_{i,c}), \\
    \mathcal{L}_{tri} = \sum_{i=1}^{B} \left[ \max(0, \, euc(\mathbf{f}^a_i, \mathbf{f}^p_i) - euc(\mathbf{f}^a_i, \mathbf{f}^n_i) + \alpha) \right], \label{eqn:triplet_loss}
\end{gather}
where $B$ is the batch size, $i$ is the instance index, $y_{i,c}$ is the one hot label of class $c$, $\hat{y}_{i,c}$ is the softmax probability score for class $c$. In equation \eqref{eqn:triplet_loss}, $euc(\mathbf{x},\mathbf{y})$ is the euclidean distance between feature $\mathbf{x}$ and $\mathbf{y}$, $\mathbf{f}^a_i$ is the anchor image feature, $\mathbf{f}^p_i$ is the feature of the positive paired instance with the maximum distance for all the positive instances in the batch, and $\mathbf{f}^n_i$ is the feature of the negative paired instance with the minimum distance for all the negative instances in the batch, and $\alpha$ is the margin for the triplet loss. 
The total ID loss is a linear combination of the classification and triplet loss with control coefficients $\lambda_c$ and $\lambda_t$:
\begin{equation}
    \mathcal{L}_{ID}=\lambda_c\mathcal{L}_{cls}+\lambda_t\mathcal{L}_{tri}.
\end{equation}

\noindent
\textbf{Overall Loss}
Our total loss is a linear combination of all the losses as formulated below, 
\begin{equation}
    \mathcal{L}_{total}=\lambda_{id}\mathcal{L}_{ID}+\lambda_b\mathcal{L}_{C_{b}}+\lambda_n\sum_{k=1}^{N}\mathcal{L}_{C_{n_k}},
\end{equation}
where $\lambda$'s are constant coefficients to control the contribution of each loss function. Without loss of generality, we set all $\lambda=1$ in all of our experiments.   

\section{Experiment}

\noindent
\textbf{Dataset and metrics} We train and evaluate our method on four CC-ReID datasets, including LTCC \cite{qian2020long}, PRCC \cite{sun2018beyond}, Celeb-reID-light \cite{huang2019celebrities} and CCVID \cite{gu2022clothes}. The LTCC contains 17,119 images of 152 identities across multiple cameras, and the PRCC comprises 33,698 images from 221 identities with two outfits each. Celeb-reID-light, a subset of Celeb-reID, includes 10,842 images of 9,021 individuals from diverse sources. CCVID, a video re-ID dataset with 2,856 sequences and 226 identities, is converted to image re-ID by using 10\% of frames for training and the first frame for testing. We evaluate model performance using Rank-1 accuracy and mean average precision (mAP) on both standard and clothing-changing re-identification tasks.


\noindent
\textbf{Implementation details} We use the EVA02-CLIP-L \cite{EVA-CLIP} model of patch size 14 as our visual encoder. Following prior work \cite{he2021transreid}, we use camera encoding as side information embedding (SIE) during positional encoding to capture the difference of viewpoints. CogVLM \cite{wang2023cogvlm} is used to extract the text description of the person in the image. We use GPT4 \cite{achiam2023gpt} to generate the language prompts for each task and summarize the biometric descriptions from the same class.  
The input image is resized to 224×224 for all datasets. The batch size is set to 8 for LTCC and 16 for other datasets. The base learning rate is set as $2\times10^{-6}$. A warming-up learning strategy with an initial learning rate of $8.42\times10^{-7}$ and the cosine learning rate decay are used to train 60 epochs. The SGD optimizer is employed in the optimization process. Moreover, the weight decay for the experiment is $0.05$. Two Nvidia Quadro RTX 5000 GPUs are used for training and testing. 

For loss function, DIFFER uses ID loss, biometric contrastive loss, and non-biometric contrastive loss in all the experiments unless specified. For the selection of non-biometric factors, the specific factors chosen for each dataset are as follows: in LTCC and PRCC, hair and clothing are used; Celeb-reID-L uses hair and pose, while CCVID focuses on clothing alone. These selections help ensure that DIFFER effectively disentangles relevant non-biometric features across varied dataset conditions. 
\YSR{can we change this to Results and comparisons? also we should start with comparison with baseline and then later compare with sota...}
\YSR{also it will be good to mention that most methods are cnn based, but we do compare with recent transformer-based methods and outperform those...}

\subsection{Comparison with the baseline method} We compare DIFFER with a baseline method that uses the same backbone architecture but is trained solely with ID loss, i.e.  DIFFER without the NBDetach module. We evaluate the performance on four benchmark datasets: LTCC, PRCC, CCVID under cloth changing setting and Celeb-reID-light in \cref{tab:comparison_baseline}. For LTCC, DIFFER achieves a top-1 accuracy improvement from 54.6\% to 58.2\% and an increase in mAP from 30.4\% to 31.6\%. Similarly, in PRCC, DIFFER raises top-1 accuracy from 65.1\% to 68.5\% and mAP from 63.0\% to 64.7\%. Celeb-reID-light and CCVID also show notable improvements, with top-1 and mAP gains for both datasets. These results highlight its effectiveness DIFFER in enhancing re-identification performance. 

\begin{table}[t]
    \centering
    \resizebox{\columnwidth}{!}{ 
    \begin{tabular}{c|c c|c c|c c|c c}
         \hline
         \multirow{2}{*}{Method} & \multicolumn{2}{c|}{\textbf{LTCC}} & \multicolumn{2}{c|}{\textbf{PRCC}} & \multicolumn{2}{c|}{\textbf{CCVID}} & \multicolumn{2}{c}{\textbf{Celeb-L}} \\
          \cline{2-9}
         & \multicolumn{1}{c}{top1} & \multicolumn{1}{c|}{mAP} & \multicolumn{1}{c}{top1} & \multicolumn{1}{c|}{mAP} & \multicolumn{1}{c}{top1} & \multicolumn{1}{c|}{mAP} & \multicolumn{1}{c}{top1} & \multicolumn{1}{c}{mAP} \\
         \hline
         Baseline & 54.6 & 30.4 & 65.1 & 63.0 & 86.9 & 86.7 &72.7 & 53.3   \\
         DIFFER & \textbf{58.2} & \textbf{31.6} & \textbf{68.5} & \textbf{64.7}  & \textbf{87.9} & \textbf{87.4}  & \textbf{75.6} & \textbf{54.3}  \\
         \hline
    \end{tabular}
    }
    \caption{Comparison DIFFER with baseline method across different datasets. The LCTT, PRCC, and CCVID are all evaluated under the clothes-changing setting. Celeb-L stands for Celeb-reID-light dataset.}
    \label{tab:comparison_baseline}
\end{table}

\subsection{Ablations and analysis}
We perform ablations on different model architectures, loss terms, VLMs, and varying non-biometric factors, along with subspace dimensions. 

\subsubsection{Model Architectures}
\cref{tab:modelArchitecture} presents the performance comparison between the baseline method and our proposed DIFFER approach across four model architectures: ClipViT-B/16, ClipViT-L/14\cite{radford2021learning}, EVA2-CLIP-B/16, and EVA2-CLIP-L/14\cite{EVA-CLIP}. We evaluate the models on LTCC under two settings. Across all model configurations, DIFFER consistently outperforms the baseline methods. DIFFER can increase the CLIPViT model top1 accuracy by 2\% and more than 3\% under the CC setting. The EVA2-CLIP-L model with DIFFER yields the highest performance, demonstrating the strength of our approach when combined with larger models. This overall trend indicates that DIFFER is highly effective at enhancing CC-ReID performance across varied model architectures.
\begin{table}[t]
    \centering
    \small
    \begin{tabular}{cc|cc|cc}
    \hline
         \multirow{2}{*}{Model} &  \multirow{2}{*}{Method} &  \multicolumn{2}{c|}{CC}&  \multicolumn{2}{c}{General}\\
         \cline{3-6}
         &   &top1&  mAP&  top1& mAP\\
         \hline
          \multirow{2}{*}{ClipViT-B}&   Baseline&34.2&  15.0&  67.5& 31.9\\
         &   DIFFER&36.5&  15.6&  72.6& 35.2\\
         \hline
          \multirow{2}{*}{ClipViT-L}&   Baseline&42.6&  19.5&  75.7& 38.0\\
         &   DIFFER&44.4&  20.0&  78.3& 39.7\\
         \hline
          \multirow{2}{*}{EVA2-CLIP-B}&   Baseline&37.5&  18.3&  76.9& 40.4\\
         &   DIFFER&40.6&  21.7&  74.0& 41.0\\
         \hline
          \multirow{2}{*}{EVA2-CLIP-L}&   Baseline&54.6&  31.2&  81.9& 50.4\\
         &   DIFFER&\textbf{58.2}&  \textbf{31.6}&  \textbf{85.0}& \textbf{52.8}\\
         \hline
        \end{tabular}
    \caption{Different model architectures performance on LTCC under both CC and general setting. }
    \label{tab:modelArchitecture}
    
\end{table}

\subsubsection{Loss functions}

We study the effect of different loss functions on PRCC and LTCC datasets under the cloth-changing setting (\cref{tab:loss}). The loss functions include identity loss $\mathcal{L}_{ID}$, biometric contrastive loss $\mathcal{L}_{C_{b}}$, and non-biometric loss $\mathcal{L}_{C_{n}}$. For the non-biometric loss, hair and clothing attributes are used. As shown in the table, both biometric loss and non-biometric loss improve the performance on both datasets, and the combination of both losses achieves the best accuracy, improving the Rank-1 accuracy from $54.6\%$ to $58.2\%$ on LTCC dataset and, $65.1\%$ to $68.5\%$ on PRCC dataset.
\begin{table}[t]
    \centering
    \small
    \begin{tabular}{ccc|cc|cc}
    \hline
          \multirow{2}{*}{$\mathcal{L}_{ID}$} &   \multirow{2}{*}{$\mathcal{L}_{C_{b}}$}&   \multirow{2}{*}{$\mathcal{L}_{C_{n}}$}&  \multicolumn{2}{c|}{LTCC}&  \multicolumn{2}{c}{PRCC}  \\
           \cline{4-7}
         &  &  &  top1&  mAP&  top1& mAP\\
         \hline
          \checkmark &  &  &  54.6&  30.4&  65.1& 63.0\\
          \checkmark &   \checkmark &  &  55.4&  30.8& 66.7& 63.6\\
          \checkmark &  &   \checkmark&  55.6&  31.0&  67.2& 63.8\\
          \checkmark &   \checkmark &   \checkmark &  \textbf{58.2}&  \textbf{31.6}&  \textbf{68.5}& \textbf{64.7}\\
           \hline
    \end{tabular}
    \caption{Ablation study of loss functions. The table evaluates the impact of combining different losses on the LTCC and PRCC cloth-changing scenario.}
    \label{tab:loss}
\end{table}

\subsubsection{Single non-biometric factor}
DIFFER could use different non-biometric factors in the NBDetach module, including clothing, pose, and hairstyle. Here, we use a single non-biometric projection head to evaluate which non-biometric factor has the most negative effect on the model performance in \cref{tab:nonbio_head}. 
We perform these experiments on LTCC, PRCC, and Celeb-reID-light under the clothes-changing setting. 
As shown in \cref{tab:nonbio_head},  different non-biometric factors play various roles in different datasets. Disentangling factors like hair or cloth can significantly improve the model's performance on the LTCC and PRCC datasets, whereas hairstyle and pose play a slightly more important role in the Celeb-reID-light dataset. So the choice of non-biometric features should be dataset-specific and different selections of non-biometric factors should be used in different user cases. 

\newcolumntype{Y}{>{\centering\arraybackslash}m{0.52cm}}  

\begin{table}[t]
    \centering
    \small
    \begin{tabularx}{0.46\textwidth}{c | *{2}{X}| *{2}{X}| *{2}{X}}
    \hline
         \multirow{2}{*}{NBF} &  \multicolumn{2}{c|}{LTCC}&  \multicolumn{2}{c|}{PRCC}& \multicolumn{2}{c}{Celeb-L}\\
          \cline{2-7}
         &   top1&  mAP&  top1&  mAP& top1&mAP\\
          \hline
        None &   55.4&  30.8& 66.7& 63.6  &74.1 & 54.7   \\
         Hair &  \textbf{56.6} &  \underline{31.3} &  \textbf{68.4} & \underline{64.0} & \textbf{75.2} & \textbf{55.0}\\       
         Cloth &  \textbf{56.6} & 30.4 & \underline{67.3} &  \textbf{64.2} & 74.2&54.8\\
         Pose & 56.1 & \textbf{31.5} & 66.5 &  63.6 & \underline{75.1} & \underline{54.9} \\
         
         \hline
    \end{tabularx}
    \caption{Analysis of different non-biometric factors (NBF). 'None' indicates that only the biometric contrastive loss is used. (Celeb-L: Celeb-reID-light)
    \YSR{can we add baseline on top? without anything...}\         \LX{done}
        }
    \label{tab:nonbio_head}
    \vspace{-3mm}
\end{table}

\begin{table}[t]
    \centering
    \small
    \renewcommand{\arraystretch}{1} 
    \resizebox{0.42\textwidth}{!}{
    \begin{tabularx}{0.46\textwidth}{p{0.6cm} p{0.6cm} p{0.6cm} | *{2}{X}| *{2}{X}}
    \hline
         \multirow{2}{*}{Hair} &  \multirow{2}{*}{Cloth} & \multirow{2}{*}{Pose} &  \multicolumn{2}{c|}{1024 dim}&  \multicolumn{2}{c}{512 dim}\\
          \cline{4-7}
         &  &  &  top1&  mAP&  top1&  mAP\\
          \hline
         \checkmark &  &  &  55.8&  29.5&  56.6& 31.3\\
         &  \checkmark &  &  \textbf{57.1} & 31.4 &  56.6& 30.4\\
         &  &  \checkmark & \textbf{57.1} &  30.4& 56.1& \underline{31.5}\\
         \hline
         \checkmark &  \checkmark &  &  54.1&  \textbf{32.0}&  \textbf{58.2}& \textbf{31.6}\\
         &  \checkmark &  \checkmark &  56.1&  31.3&  56.6& \underline{31.5}\\
         \checkmark &  &  \checkmark &  56.1&  \underline{31.8} &  57.4& 31.4\\
         \hline
         \checkmark &  \checkmark &  \checkmark &  55.1&  31.0& \underline{57.7}& 31.2\\
         \hline
    \end{tabularx}
    }
    \caption{Performance comparison with multiple non-biometric factors and subspace feature dimensions. The results are evaluated on the LTCC dataset under the clothes-changing setting.
    \YSR{do we have an explanation for this? this doesn't look good... do we have to show this? we can blame it on the challenges with adversarial learning? did you try experiment where all factors are concatenated together?}
    }
    \label{tab:multipleFactor}
\end{table}

\subsubsection{Multiple non-biometric factors}
We explore whether incorporating multiple non-biometric factors could enhance the model's performance on LTCC in \cref{tab:multipleFactor}. Our results reveal that the effect of multiple non-biometric factors depends on the relationship between the image feature dimension and the biometric/non-biometric subspace dimensions. When the image feature dimension (1024) matches the subspace dimensions (1024), using multiple non-biometric factors reduces performance, with Rank-1 accuracy dropping from 57.1\% to 54.1\%. 
Conversely, when the biometric and non-biometric subspaces are smaller than the image feature dimension (512 vs. 1024), adding multiple factors enhances Rank-1 accuracy from 56.1\% to 58.2\%. While higher-dimensional textual feature spaces offer greater representational capacity, they do not necessarily facilitate effective feature disentanglement. This aligns with prior findings \cite{materzynska_disentangling_2022}, which show that optimal visual and textual feature disentanglement is achieved in lower-dimensional subspaces. Furthermore, adding all three non-biometric factors reduces performance even in the 512-dimensional subspace, likely because this dimension is too large to effectively disentangle the 1024-dimensional space into three subspaces. 

\YSR{do we have experiments with 64?} \LX{because the smallest text encoder dim is 512. so we cannot go below to 64}

\subsubsection{Experiment using different VLMs}
We conducted experiments using different VLMs, including LLama1.6 \cite{liu2023improved} 
and InternVL2.5 \cite{chen2024expanding}, in addition to CogVLM-18B in \cref{tab:vlm}. 
The results indicate that all three models show similar performance, suggesting that these large-scale visual-language models are equally good at giving low-level person appearance descriptions.
\begin{table}[t]
    \centering
    \resizebox{0.38\textwidth}{!}{ 
    \begin{tabular}{cc|cc}
        \hline
        \multicolumn{2}{c|}{VLM Model} & \multicolumn{2}{c}{LTCC} \\
        \cline{1-2} \cline{3-4}
        Model & \# Parameters & Rank-1 & mAP \\
        \hline
        LLama1.6 & 7B & 57.9 & 31.7 \\
        CogVLM & 18B & \textbf{58.2} & 31.6 \\
        InternVL2.5 & 26B & 57.7 & \textbf{32.6} \\
        \hline
    \end{tabular}
    }
    \caption{Experiment on different VLMs. The performance is evaluated on LTCC under the clothes-changing setting.}
    \label{tab:vlm}
    \vspace{-3mm}
\end{table}

\noindent

\subsection{Comparison with state-of-the-art methods}


\renewcommand{\arraystretch}{1.1}
\newcolumntype{C}[1]{>{\centering\arraybackslash}m{#1}}

\begin{table*}[ht!]
\centering
\footnotesize
\resizebox{\textwidth}{!}{
\begin{tabularx}{\textwidth}{c c C{0.14\linewidth} c| *{2}{X}| *{2}{X}| *{2}{X}| *{2}{X}}
\hline
\multirow{3}{*}{Method} & \multirow{3}{*}{Venue/Year} & \multirow{3}{*}{Modality} & \multirow{3}{*}{Backbone} & \multicolumn{4}{c|}{PRCC} & \multicolumn{4}{c}{LTCC} \\
\cline{5-12}
 &  &  &  & \multicolumn{2}{c|}{CC}  & \multicolumn{2}{c|}{SC} & \multicolumn{2}{c|}{CC} & \multicolumn{2}{c}{General} \\
 \cline{5-12}
 &  &  &  & top1 & mAP & top1 & mAP & top1 & mAP & top1 & mAP\\
\hline
3DSL\cite{chen2021learning} & CVPR 21 & RGB+pose+sil.+3D & ResNet-50x2 & - &  51.3& - & - &  31.2&  14.8 & - & -\\
GI-ReID \cite{jin2022cloth} & CVPR 22 & RGB+pose & OSNet &  -&  37.5&  -&  -&  28.11&  13.17&  63.2& 29.4\\
CAL \cite{gu2022clothes} & CVPR 22 & RGB & ResNet-50 & 55.2&  55.8&  \textbf{100}&  \underline{99.8}&  40.1&  18&  74.2& 40.8\\
AIM \cite{yang2023good} & CVPR 23 & RGB & ResNet-50x2 &  57.9&  58.3&  \textbf{100}&  \textbf{99.9}&  40.6&  19.1&  76.3& 41.1\\
LDF \cite{chan2023learning} & ACM 23 & RGB & ResNet-50 &  58.4&  58.6&  \textbf{100}&  99.7&  32.9&  15.4&  73.4&- \\
3DInvarReID \cite{liu2023learning} & ICCV 23 & RGB+3D & - &  56.5&  57.2& - & - &  40.9&  18.9& - & -\\
\rowcolor{gray!20} SCNet \cite{guo2023semantic} & ACM 23 & RGB+parsing/ RGB(Inference) & ResNet-50 & 58.7&  \underline{59.9}&  \textbf{100}&  97.8&  \underline{47.5}&  \underline{25.5}&  76.3& 43.6\\
CVSL \cite{nguyen2024contrastiveViewpoint} & WACV 24 & RGB + pose & ResNet-50 & 57.5&  56.9&  97.5&  99.1&  44.5&  21.3&  76.4& 41.9\\
\rowcolor{gray!20} CLIP3DReID \cite{liu2024distilling} & CVPR 24 & RGB+desc.+3D/ RGB(Inference) & ResNet-50 & 60.6& 59.3 & - &  - &  42.1&  21.7& - & -\\
CGPG \cite{nguyen2024contrastiveClothing} & CVPR 24 & RGB+sil. & ResNet-50 &  61.8&  58.3&  \textbf{100}&  99.6&  46.2&  22.9&  77.2& 42.9\\
\hline
TransReID\cite{he2021transreid} & CVPR 21 & RGB & ViT-B & 46.6&  44.8&  \textbf{100}&  99&  34.4&  17.1&  70.4 & 37.0 \\
Instruct-ReID\cite{he_instruct-reid_2023} & CVPR 24 & RGB + instruct & ViT-B+ALBEF & 54.2 &  52.3 &  - &  - & -&  -&  75.8& \underline{52.0}\\
MADE \cite{peng2024masked} & IEEE TMM 24 & RGB+attributes & EVA2-CLIP-L & \underline{64.3}&  59.1&  \textbf{100}&  98.6&  47.4&  24.4&  \underline{84.2}& 48.2\\
\hline
\textbf{DIFFER} &  & RGB+description/ RGB(Inference) & EVA2-CLIP-L & \textbf{68.5} & \textbf{64.7} & \underline{99.9} & 99.5 & \textbf{58.2} & \textbf{31.6} & \textbf{85.0} & \textbf{52.8} \\
\hline
\end{tabularx}}
\caption{Comparison of state-of-the-art methods on PRCC and LTCC datasets. CC stands for clothes-changing, SC stands for same-clothes, and General stands for both CC and SC. The best results are highlighted with bold and the second best results are underlined. 
\YSR{underline second best...}\LX{done}
}
\label{tab:comparisonSOTA}
\vspace{-3mm}
\end{table*}

We compare DIFFER with existing CNN-based clothing-changing person REID methods, including 
 3DSL \cite{chen2021learning}, 
GI-ReID \cite{jin2022cloth}, 
CAL \cite{gu2022clothes}, 
AIM \cite{yang2023good}, 
LDF \cite{chan2023learning}, 3DInvarReID \cite{liu2023learning}, 
SCNet \cite{guo2023semantic}, 
CVSL \cite{nguyen2024contrastiveViewpoint}, CLIP3DReID \cite{liu2024distilling}, CCPG \cite{nguyen2024contrastiveClothing},  and ViT based methods including TransReID \cite{he2021transreid}, Instruct-ReID\cite{he_instruct-reid_2023}, MADE \cite{peng2024masked}. We evaluate the performance on two datasets, PRCC and LTCC, in \cref{tab:comparisonSOTA}. 

As shown in the table, DIFFER outperforms other methods across most evaluation metrics in both datasets, benefiting from the robust capability of the multimodal foundation model and the proposed NBDetach module. For the PRCC dataset, DIFFER achieves the highest top-1 accuracy (68.5\%) and mAP (64.7\%) in the cloth-changing scenario; in the LTCC dataset, DIFFER again leads with top-1 accuracy of 58.2\% and mAP of 31.6\% in the cloth-changing scenario, demonstrating its effectiveness in long-term clothes changing re-identification tasks. 


\begin{figure}[t!]
\centering
\includegraphics[width=7.5cm]{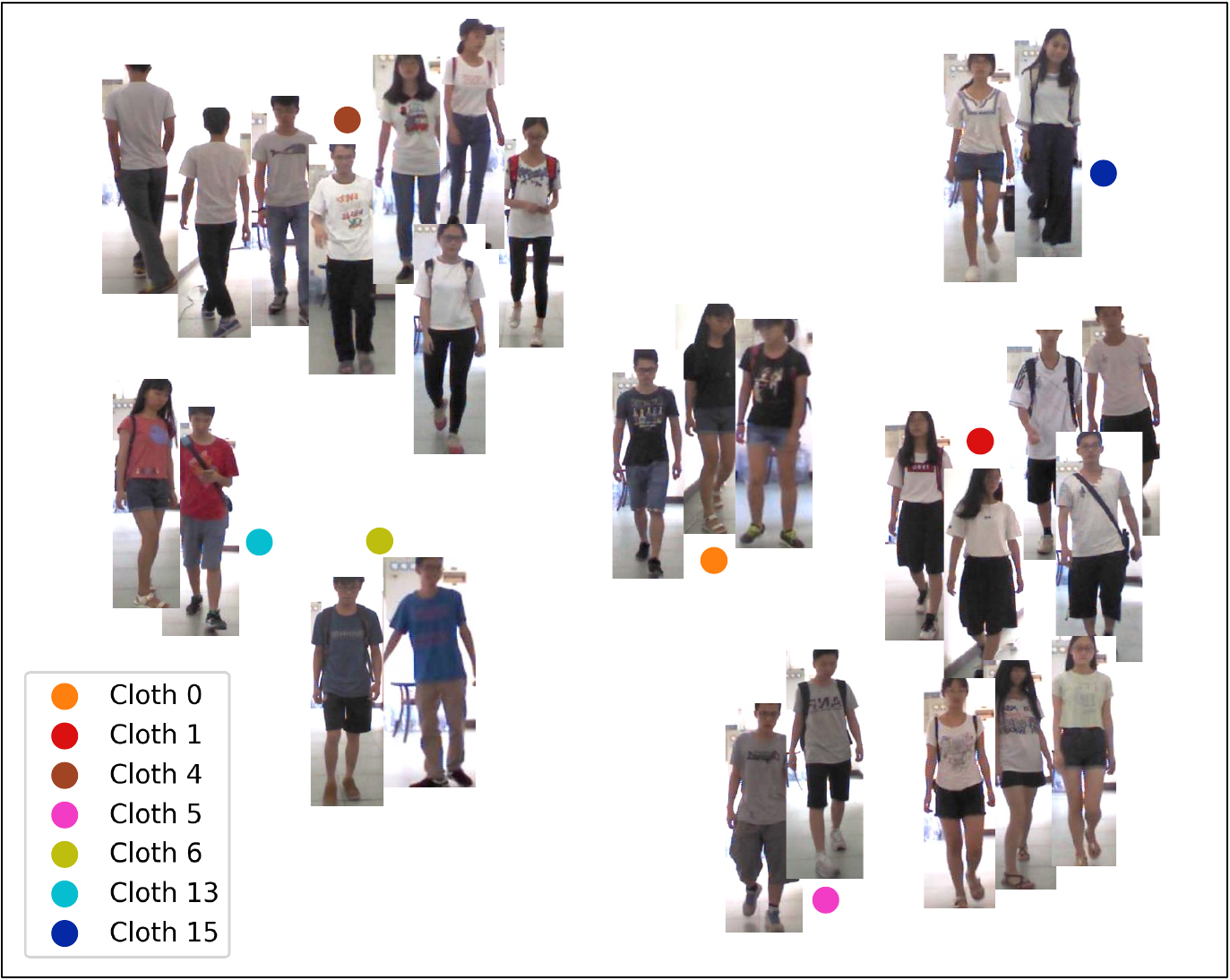}
\caption{Clothing textual feature cluster visualization results. We visualize the cluster results of different images based on the clothing textual feature. As shown in the figure, images with similar outfits are successfully grouped together. 
}

  \label{fig:clusterCloth}
   \vspace{-3mm}
\end{figure}

\subsection{Effectiveness of textual features} To demonstrate the effectiveness of the textual features, we visualize the textual feature clustering results in \cref{fig:clusterCloth}. We use the clothes textual feature for PRCC datasets and cluster the clothing feature with the DBSCAN \cite{ester1996density} cluster algorithm to label different clothes with different IDs. Then t-SNE \cite{van2008visualizing} is used to visualize the cluster results and we overlay the corresponding images on top of the feature representation. As shown in \cref{fig:clusterCloth}, the clothing textual feature successfully clusters the outfits in the image regardless of gender or body shape. The results demonstrate that our clothing textual features could measure the outfit similarity between different people, while previous clothes labels only focused on differentiating the clothes for one individual. 

\begin{figure}[t]
    \centering
    \setlength{\fboxrule}{1pt} 
    \setlength{\fboxsep}{0pt}

     \vspace{0.15cm}
    \begin{subfigure}{0.036\textwidth}
        \centering      
        \includegraphics[width=0.9\textwidth]{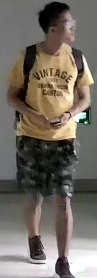}
    \end{subfigure}
    \hspace{0.02\textwidth}
     \begin{subfigure}{0.036\textwidth}
        \centering
        \fcolorbox{red}{white}{\includegraphics[width=0.9\textwidth]{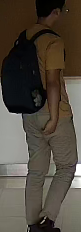}}
    \end{subfigure}
     \hfill
    \begin{subfigure}{0.036\textwidth}
        \centering
        \fcolorbox{red}{white}{\includegraphics[width=0.9\textwidth]{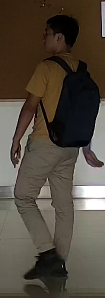}}
    \end{subfigure}
     \hfill
    \begin{subfigure}{0.036\textwidth}
        \centering
        \fcolorbox{red}{white}{\includegraphics[width=0.9\textwidth]{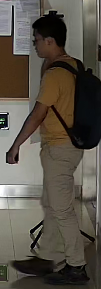}}
    \end{subfigure}
    \hfill
    \begin{subfigure}{0.036\textwidth}
        \centering
        \fcolorbox{red}{red}{\includegraphics[width=0.9\textwidth]{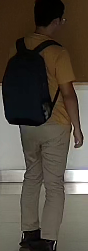}}
    \end{subfigure}
    \hfill
    \begin{subfigure}{0.036\textwidth}
        \centering
       \fcolorbox{red}{white}{\includegraphics[width=0.9\textwidth]{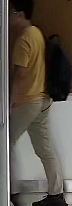}}
    \end{subfigure}
    \begin{subfigure}{0.036\textwidth}
        \centering
        \fcolorbox{red}{white}{\includegraphics[width=0.9\textwidth]{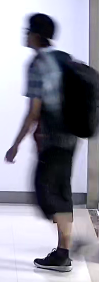}}
    \end{subfigure}
     \hfill
    \begin{subfigure}{0.036\textwidth}
        \centering
        \fcolorbox{red}{white}{\includegraphics[width=0.9\textwidth]{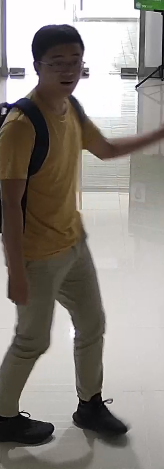}}
    \end{subfigure}
     \hfill
    \begin{subfigure}{0.036\textwidth}
        \centering
        \fcolorbox{red}{white}{\includegraphics[width=0.9\textwidth]{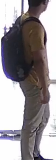}}
    \end{subfigure}
    \hfill
    \begin{subfigure}{0.036\textwidth}
        \centering
        \fcolorbox{red}{red}{\includegraphics[width=0.9\textwidth]{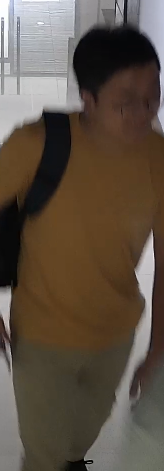}}
    \end{subfigure}
    \hfill
    \begin{subfigure}{0.036\textwidth}
        \centering
       \fcolorbox{red}{white}{\includegraphics[width=0.9\textwidth]{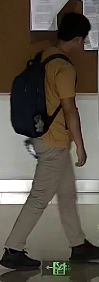}}
    \end{subfigure}

     \vspace{0.05cm}
    \begin{subfigure}{0.036\textwidth}
        \centering      
         \fcolorbox{white}{white}{\rule{0pt}{0cm}} 
    \end{subfigure}
    \hspace{0.02\textwidth}
     \begin{subfigure}{0.036\textwidth}
        \centering
        \fcolorbox{blue}{white}{\includegraphics[width=0.9\textwidth]{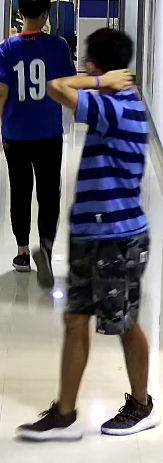}}
    \end{subfigure}
     \hfill
    \begin{subfigure}{0.036\textwidth}
        \centering
        \fcolorbox{blue}{white}{\includegraphics[width=0.9\textwidth]{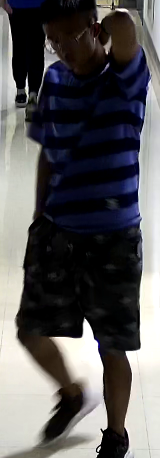}}
    \end{subfigure}
     \hfill
    \begin{subfigure}{0.036\textwidth}
        \centering
        \fcolorbox{blue}{white}{\includegraphics[width=0.9\textwidth]{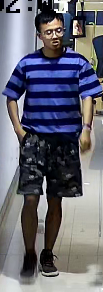}}
    \end{subfigure}
    \hfill
    \begin{subfigure}{0.036\textwidth}
        \centering
        \fcolorbox{blue}{white}{\includegraphics[width=0.9\textwidth]{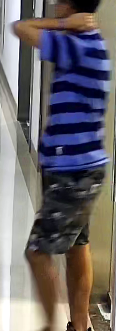}}
    \end{subfigure}
    \hfill
    \begin{subfigure}{0.036\textwidth}
        \centering
       \fcolorbox{blue}{white}{\includegraphics[width=0.9\textwidth]{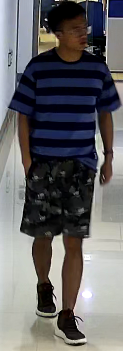}}
    \end{subfigure}
    \begin{subfigure}{0.036\textwidth}
        \centering
        \fcolorbox{red}{white}{\includegraphics[width=0.9\textwidth]{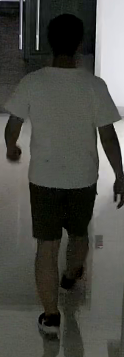}}
    \end{subfigure}
     \hfill
    \begin{subfigure}{0.036\textwidth}
        \centering
        \fcolorbox{blue}{white}{\includegraphics[width=0.9\textwidth]{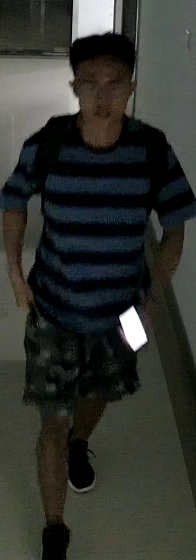}}
    \end{subfigure}
     \hfill
    \begin{subfigure}{0.036\textwidth}
        \centering
        \fcolorbox{blue}{white}{\includegraphics[width=0.9\textwidth]{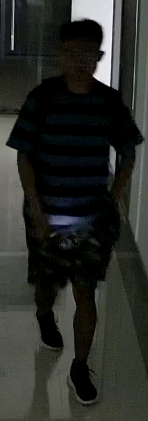}}
    \end{subfigure}
    \hfill
    \begin{subfigure}{0.036\textwidth}
        \centering
        \fcolorbox{red}{white}{\includegraphics[width=0.9\textwidth]{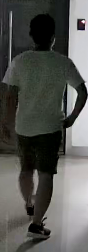}}
    \end{subfigure}
    \hfill
    \begin{subfigure}{0.036\textwidth}
        \centering
       \fcolorbox{red}{white}{\includegraphics[width=0.9\textwidth]{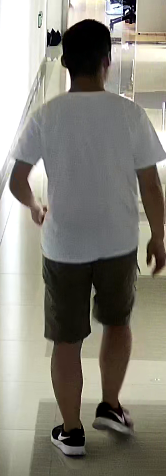}}
    \end{subfigure}

    \vspace{0.15cm}

    \begin{subfigure}{0.036\textwidth}
        \centering      
        \includegraphics[width=0.9\textwidth]{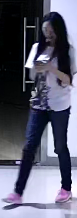}
    \end{subfigure}
    \hspace{0.02\textwidth}
     \begin{subfigure}{0.036\textwidth}
        \centering
        \fcolorbox{red}{white}{\includegraphics[width=0.9\textwidth]{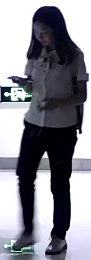}}
    \end{subfigure}
     \hfill
    \begin{subfigure}{0.036\textwidth}
        \centering
        \fcolorbox{blue}{white}{\includegraphics[width=0.9\textwidth]{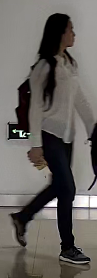}}
    \end{subfigure}
     \hfill
    \begin{subfigure}{0.036\textwidth}
        \centering
        \fcolorbox{blue}{white}{\includegraphics[width=0.9\textwidth]{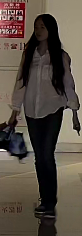}}
    \end{subfigure}
    \hfill
    \begin{subfigure}{0.036\textwidth}
        \centering
        \fcolorbox{red}{red}{\includegraphics[width=0.9\textwidth]{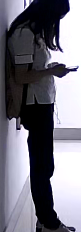}}
    \end{subfigure}
    \hfill
    \begin{subfigure}{0.036\textwidth}
        \centering
       \fcolorbox{red}{white}{\includegraphics[width=0.9\textwidth]{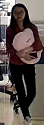}}
    \end{subfigure}
    \begin{subfigure}{0.036\textwidth}
        \centering
        \fcolorbox{blue}{white}{\includegraphics[width=0.9\textwidth]{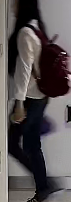}}
    \end{subfigure}
     \hfill
    \begin{subfigure}{0.036\textwidth}
        \centering
        \fcolorbox{red}{white}{\includegraphics[width=0.9\textwidth]{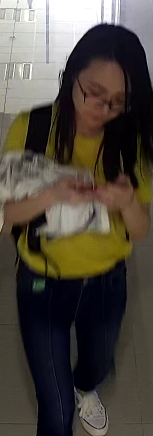}}
    \end{subfigure}
     \hfill
    \begin{subfigure}{0.036\textwidth}
        \centering
        \fcolorbox{red}{white}{\includegraphics[width=0.9\textwidth]{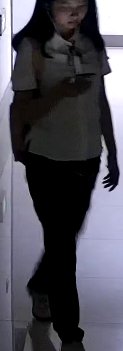}}
    \end{subfigure}
    \hfill
    \begin{subfigure}{0.036\textwidth}
        \centering
        \fcolorbox{red}{red}{\includegraphics[width=0.9\textwidth]{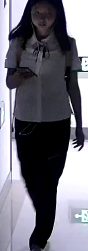}}
    \end{subfigure}
    \hfill
    \begin{subfigure}{0.036\textwidth}
        \centering
       \fcolorbox{red}{white}{\includegraphics[width=0.9\textwidth]{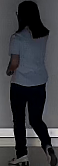}}
    \end{subfigure}

    \vspace{0.05cm}
     \begin{subfigure}{0.036\textwidth}
        \centering      
       \fcolorbox{white}{white}{\rule{0pt}{0cm}} 
        \caption*{}
    \end{subfigure}
    \hspace{0.02\textwidth}
     \begin{subfigure}{0.036\textwidth}
        \centering
        \fcolorbox{blue}{white}{\includegraphics[width=0.9\textwidth]{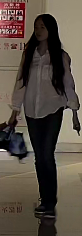}}
    \end{subfigure}
    \hfill
    \begin{subfigure}{0.036\textwidth}
        \centering
       \fcolorbox{red}{white}{\includegraphics[width=0.9\textwidth]{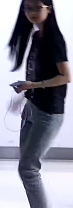}}
    \end{subfigure}
     \hfill
    \begin{subfigure}{0.036\textwidth}
        \centering
        \fcolorbox{blue}{white}{\includegraphics[width=0.9\textwidth]{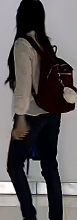}}
    \end{subfigure}
     \hfill
    \begin{subfigure}{0.036\textwidth}
        \centering
        \fcolorbox{red}{white}{\includegraphics[width=0.9\textwidth]{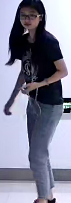}}
    \end{subfigure}
    \hfill
    \begin{subfigure}{0.036\textwidth}
        \centering
        \fcolorbox{blue}{red}{\includegraphics[width=0.9\textwidth]{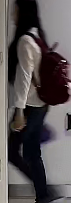}}
    \end{subfigure}
    \hfill
    \begin{subfigure}{0.036\textwidth}
        \centering
       \fcolorbox{red}{white}{\includegraphics[width=0.9\textwidth]{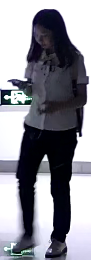}}
    \end{subfigure}
    \begin{subfigure}{0.036\textwidth}
        \centering
        \fcolorbox{red}{white}{\includegraphics[width=0.9\textwidth]{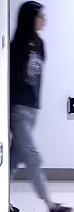}}
    \end{subfigure}
     \hfill
    \begin{subfigure}{0.036\textwidth}
        \centering
        \fcolorbox{blue}{white}{\includegraphics[width=0.9\textwidth]{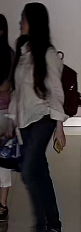}}
    \end{subfigure}
     \hfill
    \begin{subfigure}{0.036\textwidth}
        \centering
        \fcolorbox{red}{white}{\includegraphics[width=0.9\textwidth]{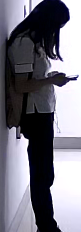}}
    \end{subfigure}
    \hfill
    \begin{subfigure}{0.036\textwidth}
        \centering
        \fcolorbox{red}{white}{\includegraphics[width=0.9\textwidth]{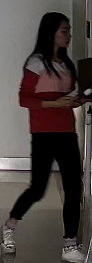}}
    \end{subfigure}

    \caption{The Top-10 retrieval results from LTCC dataset with baseline and DIFFER. Each row displays the ranked results for a query image (left), followed by Rank-1 to Rank-10 retrieval images from left to right. The first row shows baseline results, and the second shows DIFFER results, with correct matches in blue boxes and incorrect matches in red.
\YSR{it looks little off, maybe the boundaries are too wide... see prior works, how they do it...}
\YSR{also for caption of figures/tables, see fig 1 and 2... follow that... start with short description with bold italics, what it is at a high level...}
\YSR{green for correct was good... blue is fine too...maybe you can show top 10, one comment, space between images from two rows should be less then images from same row, for good grouping ... adding top 10 will help... we can use this for error analysis...}
    }
    \label{fig:topkMatch}
    \vspace{-3mm}
\end{figure}

\subsection{Retrieval result visualization} 
We visualize the top 10 matched gallery images for some query images from LTCC in \cref{fig:topkMatch}. As illustrated, our model successfully identifies individuals despite variations in clothing and pose, while the baseline method struggles to differentiate between individuals with similar clothing and hairstyles, even when body shapes differ. However, we observe that DIFFER encounters large incorrect matches in the second sample, where most retrieved results share similar body shapes, hairstyles, and occlusion. This indicates the model still faces challenges with occlusions and visual similarities in body angle and shape. Incorporating finer-grained biometric features could improve its ability to distinguish between visually similar individuals.

\YSR{it will be good to discuss some failure cases from the retreived images... just see it should not harm ... but we might see some interesting falure cases which are hard...}

\vspace{-1mm}

\section{Conclusion}
In this work, we propose DIFFER, a novel method for clothes-changing person re-identification that disentangles biometric and non-biometric features using a multimodal approach. By leveraging pseudo-labels generated from large visual-language models, DIFFER effectively separates identity-related information from non-biometric attributes. Unlike many existing approaches, DIFFER does not rely on additional external modalities during inference, making it more reliable and flexible. Our extensive evaluation on multiple benchmark datasets demonstrates that DIFFER consistently outperforms the baseline method. 

\vspace{-3mm}

\paragraph{Limitations} While DIFFER shows strong performance in disentangling non-biometric features, limitations remain. Combining multiple non-biometric aspects can destabilize the gradient reversal process, but future advancements in large VLMs with extended context lengths may enable us to input all non-biometric descriptions simultaneously, enhancing stability. We also recognize the ethical implications of person re-identification (ReID), as misuse could lead to privacy violations. We are committed to responsible development and align with the ReID community in advocating for safeguards against misuse.
\YSR{we can add few sentences on limitations... - combining different non-bio aspects, if we use lvms with bigger context length, we can potentially pass all non-bio descriptions at once and the gr will be more stable... something like this.. or think of something else... at the end also say, this is reid and with wrong intentions it can be harmful too... we will try our best to make sure that do not happen... just see prior reid works if they have some wordings for this...}

\noindent

{
    \small
    \bibliographystyle{ieeenat_fullname}

}

\setcounter{section}{0}
\renewcommand{\thesection}{S\arabic{section}}
\setcounter{figure}{0}
\renewcommand{\thefigure}{S\arabic{figure}}
\setcounter{table}{0}
\renewcommand{\thetable}{S\arabic{table}}

\clearpage
\setcounter{page}{1}
\maketitlesupplementary

In this supplementary, we include details that we were not able to provide in the main paper due to space constraints. It includes,

\begin{itemize}
    \item Details on image text descriptions with examples
    \item Description summarization prompt and experiment details
    \item Implementation details of the NBDetach module
    \item Disentangled identity Feature visualization
    \item Examples of enhanced performance on baseline failure cases.
\end{itemize}

\section{Image Text Description Example} \label{vlmExample}

\begin{wrapfigure}{R}{0.1\textwidth}
    \centering
    \includegraphics[width=0.1\textwidth]{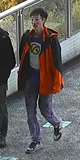}
\end{wrapfigure}

The following is an example of a conversation between the user and the visual-language model CogVLM \cite{wang2023cogvlm}. The corresponding input image is displayed to the left of the passage \YSR{where? i see one on right? maybe just add figure number as refer to it...}. The description of biometric characteristics includes details about the individual’s physical attributes, such as age, gender, height, and body type. Non-biometric factors encompass elements like hairstyle, clothing, posture, and background. As demonstrated in the example, the VLM accurately extracts the essential information required for our task but also introduces some errors. For instance, hairstyle information is incorrectly classified as a biometric attribute, and the person’s jacket is described as having black and white patches, though the white patches are actually part of the shirt underneath.

\begin{itemize}
    \item \textbf{Input}: \textit{Describe the individual's overall \textbf{physical appearance}, including estimated age, gender, height (e.g., short, average, tall based on surroundings if applicable), and build(e.g., slender, average, robust).}
    \item \textbf{CogVLM}:\textit{ The individual appears to be a young male, possibly in his late teens to early twenties. He has short, dark hair and glasses. Based on the surroundings, he seems to be of average height and build.}

    \item \textbf{Input}: \textit{Provide a detailed description of the \textbf{hair features}, including hair color, style, and length}
    \item \textbf{CogVLM}: \textit{The individual has short, dark hair. }

    \item \textbf{Human}: \textit{Detail the type of \textbf{clothing} the person is wearing(the style, colors, and any visible logos or patterns), shoes and any accessories (e.g., glasses, watches, jewelry).}
    \item \textbf{CogVLM}: \textit{The individual is wearing a red jacket with black and white patches. Underneath the jacket, he has a blue shirt with a white logo or emblem on it. He is also wearing purple pants and black and white shoes. He is wearing glasses.}

    \item \textbf{Input}: \textit{Describe the person’s \textbf{posture} when the image was taken (e.g., standing straight, leaning, walking). Note any characteristics of the gait, such as limping, brisk walking, or any peculiarities that stand out. Mention the alignment and demeanor suggested by the posture (e.g., confident, tired, hurried).}
    \item \textbf{CogVLM}: \textit{The person appears to be walking, with a somewhat brisk gait. His posture is upright, suggesting confidence, and he seems to be moving forward purposefully.}

    \item \textbf{Input}: \textit{Describe the setting or \textbf{background} in which the person is located (e.g., urban street, office, park). Identify any objects or elements in the vicinity that the person is interacting with or that are relevant to the scene. Assess the general atmosphere or mood of the environment, such as busy, tranquil, chaotic, etc.}
    \item \textbf{CogVLM}: \textit{The person appears to be in an urban setting, possibly a street or a pedestrian area. There are glass railings and a metal structure visible in the background, suggesting a public space or a walkway. The general atmosphere seems to be calm and quiet.}
\end{itemize}

\section{Description Summarization}
We use GPT-4 \cite{achiam2023gpt} to summarize all the biometric descriptions from one person to one prompt. Additionally, we conduct experiments to summarize all clothing descriptions for a specific clothing class into a unified clothing text description, ensuring that all images sharing the same clothing label have identical clothing text features. However, it is important to note that, except for this section, the results presented in this paper do not use summarized clothing descriptions to avoid reliance on extra clothing labels.
\subsection{Summary prompts}
The following are the summary prompts used to generate biometric and clothing descriptions. These summary prompts were initially created using GPT-4 and refined manually to align with our specific requirements better. 
\begin{itemize}
\item \textit{Summarize the individual's overall \textbf{physical appearance}, only including estimated age, gender, height (e.g., short, average, tall based on surroundings if applicable), and build (e.g., slender, average, robust) based on the following information. Do not summarize the hairstyle. Only include the information that most sentences agree on.}
\item \textit{Summarize the type of \textbf{clothing} the person is wearing(the style, colors, and any visible logos or patterns), shoes and any accessories (e.g., glasses, watches, jewelry) based on the following information. Using three to four describing sentences. Only include the information that most sentences agree on.}
\end{itemize}

\subsection{Summary Description Example}
\begin{figure}[t!]
    \centering
    \begin{subfigure}{0.05\textwidth}
        \centering
        \includegraphics[height=3.5cm]{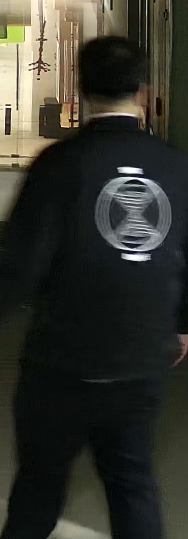}
    \end{subfigure}
    \hspace{0.5cm}
    \begin{subfigure}{0.05\textwidth}
        \centering
        \includegraphics[height=3.5cm]{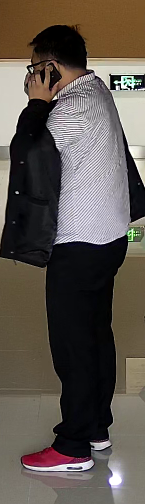}
    \end{subfigure}
    \hspace{1.2cm}
    \begin{subfigure}{0.05\textwidth}
        \centering
        \includegraphics[height=3.5cm]{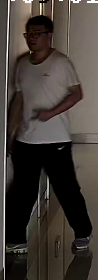}
    \end{subfigure}
    \hspace{0.5cm}
    \begin{subfigure}{0.05\textwidth}
        \centering
        \includegraphics[height=3.5cm]{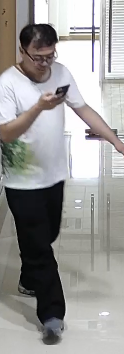}
    \end{subfigure}
    \caption{\textbf{Input images for the summary descriptions}. The four images, all representing the same individual, are used for summarizing the biometrics and clothing descriptions. The first two images are examples of the person wearing \textbf{Cloth1}, and the second two images are for \textbf{Cloth2}.}
    \label{fig:summerizepeopleExample}
\end{figure}

Here, we present examples of biometric and clothing description summaries, with corresponding example images displayed in \cref{fig:summerizepeopleExample}. The below text examples illustrate that biometric summarization effectively captures the primary physical traits of the individual, though the descriptions are generally broad and lack fine-grained specificity. In contrast, clothing summaries provide more specific details, such as the red shoes and white patterns on the jacket for Cloth1, along with the overall style. However, these summaries may unintentionally incorporate details from other images that are not visually present in the current image, introducing extraneous information. In conclusion, while biometric summaries enhance consistency and robustness in capturing identity-related features, clothing summaries may lead to inaccuracies by including details not relevant to the image being analyzed.

\begin{itemize}
    \item \textbf{Biometrics summary description}: \textit{The individual appears to be a male, primarily estimated to be in his late 20s to early 30s, with a consensus also leaning towards late teens to early 20s in several descriptions. He consistently has a medium or average build and is of average height based on the surroundings.}
     \item \textbf{Clothing summary description for Cloth1}: \textit{The individual is dressed in a casual style, predominantly featuring black clothing, including a jacket with a white logo on the back and black pants. The jacket's logo is described as either a stylized sand timer or circular. Red shoes or sneakers add a pop of color to the otherwise monochrome outfit. While there is mention of glasses being worn in two descriptions, the presence of other accessories like watches or jewelry is either not mentioned or stated to be absent.}
    \item \textbf{Clothing summary description for Cloth2}: \textit{The individual is dressed in a casual style, wearing a white t-shirt that features a small, greenish logo or design on the left side. They are also wearing black pants, complemented by gray sports shoes. Additionally, the person is accessorized with glasses, adding a practical element to their look.}
   
\end{itemize}

\subsection{Additional experiment results}
We study the effect of using summary text descriptions during training. As shown in \cref{tab:summeryTable}, the biometric summary caption could increase the accuracy from  57.7\% to 58.2\% on LTCC, from 67.3\% to 68.5\% on PRCC. On the other hand, adding the clothes summary description could decrease the performance by 0.8\% on LTCC and 1\% on PRCC. The summarized biometric descriptions likely help the model capture missing identity information, especially in cases of blurred or occluded images. Conversely, clothing descriptions depend heavily on the specific image, and summarized descriptions may introduce irrelevant information or overlook critical details, negatively affecting disentanglement performance.

\begin{table}[t!]
    \centering
    \begin{tabularx}{0.45\textwidth}{cc| *{2}{X}| *{2}{X}}
    \hline
   \multirow{2}{*}{Bio} & \multirow{2}{*}{Cloth} &  \multicolumn{2}{c|}{LTCC}&  \multicolumn{2}{c}{PRCC}\\
   \cline{3-6}
         &  & top1&  mAP & top1&  mAP \\
   \hline
      Image & Image  &   57.7& 31.3&67.3&63.8\\
 
     Summary & Image    &  \textbf{58.2} & \textbf{31.6}&\textbf{68.5} &\textbf{64.7} \\
    
    Summary & Summary  &  57.4& 30.5&67.5&\textbf{64.7}\\
     \hline
    \end{tabularx}
    \caption{\textbf{Image description and summarized description experiment results}. We compare the results of whether to use the image or summarized description for biometric contrastive loss $\mathcal{L}_{C_b}$ and clothing non-biometric contrastive loss $\mathcal{L}_{C_n}$ in the table. LTCC and PRCC datasets under the cloth-changing setting are used.
    }
    \label{tab:summeryTable}
\end{table}

\section{NBDetach Module Architecture Details}

We present a comprehensive description of the architecture of our NBDetach module. For both the biometric and non-biometric projection heads, \( H_b \) and \( H_n \), we employ linear transformations with matrix multiplication and bias addition to execute the projection. 

The input image feature \( \mathbf{f}^i \) has a dimensionality of 1024 for the EVA2-CLIP-L model. The projected biometric and non-biometric image features, \( \mathbf{f}^i_b \) and \( \mathbf{f}^i_n \), are designed to match the dimensions of their corresponding textual features \( \mathbf{f}^t_b \) and \( \mathbf{f}^t_n \), which can be 512, 768  or 1024 dimensions. In our experiments, we use the 512-dimensional textual feature for LTCC and 768-dimensional for other datasets. 
In the gradient reversal layer (GRL), we set the negative scalar \( \alpha \) to -1, which changes the gradient sign to negative, i.e. multiplies it by -1.

\setlength{\fboxsep}{1pt}

\section{Feature visualization for disentangled identity} 
\begin{figure}[t!]
    \centering
    \begin{subfigure}{0.05\textwidth}
    \centering
        \includegraphics[height=3cm]{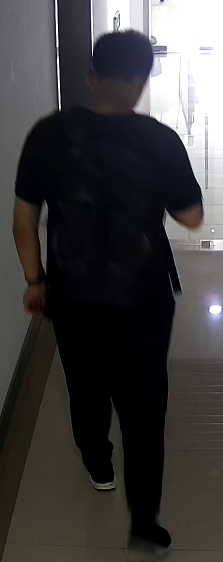}
    \end{subfigure}
    \hspace{\fill}
    \begin{subfigure}{0.05\textwidth}
        \centering
        \includegraphics[height=3cm]{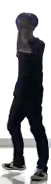}
    \end{subfigure}\hfill
    \begin{subfigure}{0.05\textwidth}
        \centering
        \includegraphics[height=3cm]{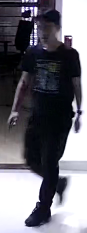}
    \end{subfigure}
    \hspace{\fill}
    \begin{subfigure}{0.05\textwidth}
        \centering
        \includegraphics[height=3cm]{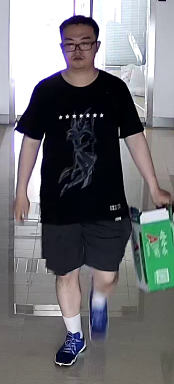}
    \end{subfigure}
    \hspace{\fill}
    \begin{subfigure}{0.05\textwidth}
        \centering
        \includegraphics[height=3cm]{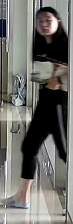}
    \end{subfigure}
    \hspace{\fill}
    \begin{subfigure}{0.05\textwidth}
        \centering
        \includegraphics[height=3cm]{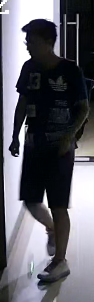}
    \end{subfigure} 
    \begin{subfigure}{0.23\textwidth}
        \centering
        \includegraphics[width=0.95\textwidth]{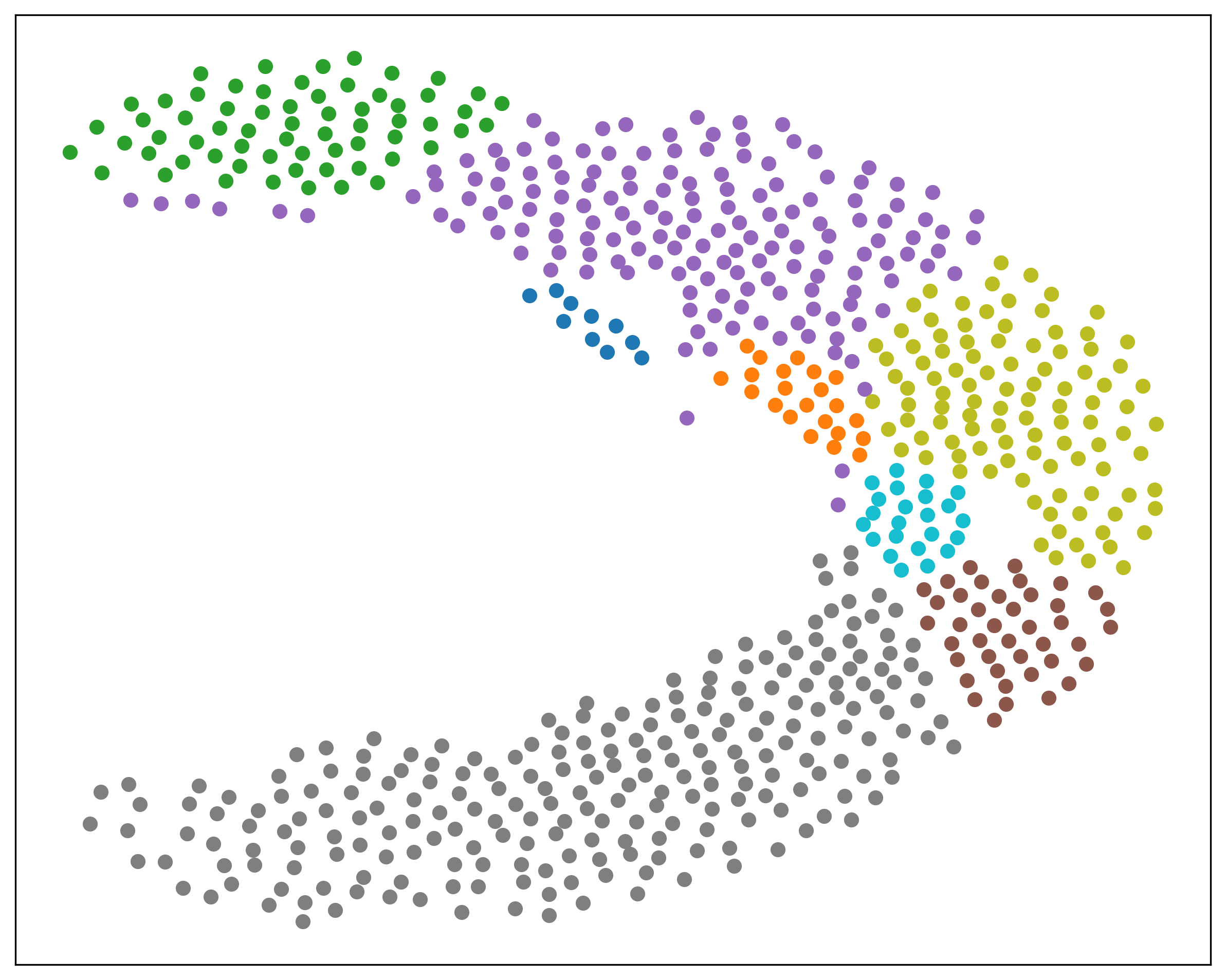}
        \label{fig:image1}
    \end{subfigure}
    \hspace{\fill}
    \begin{subfigure}{0.23\textwidth}
        \centering
        \includegraphics[width=0.95\textwidth]{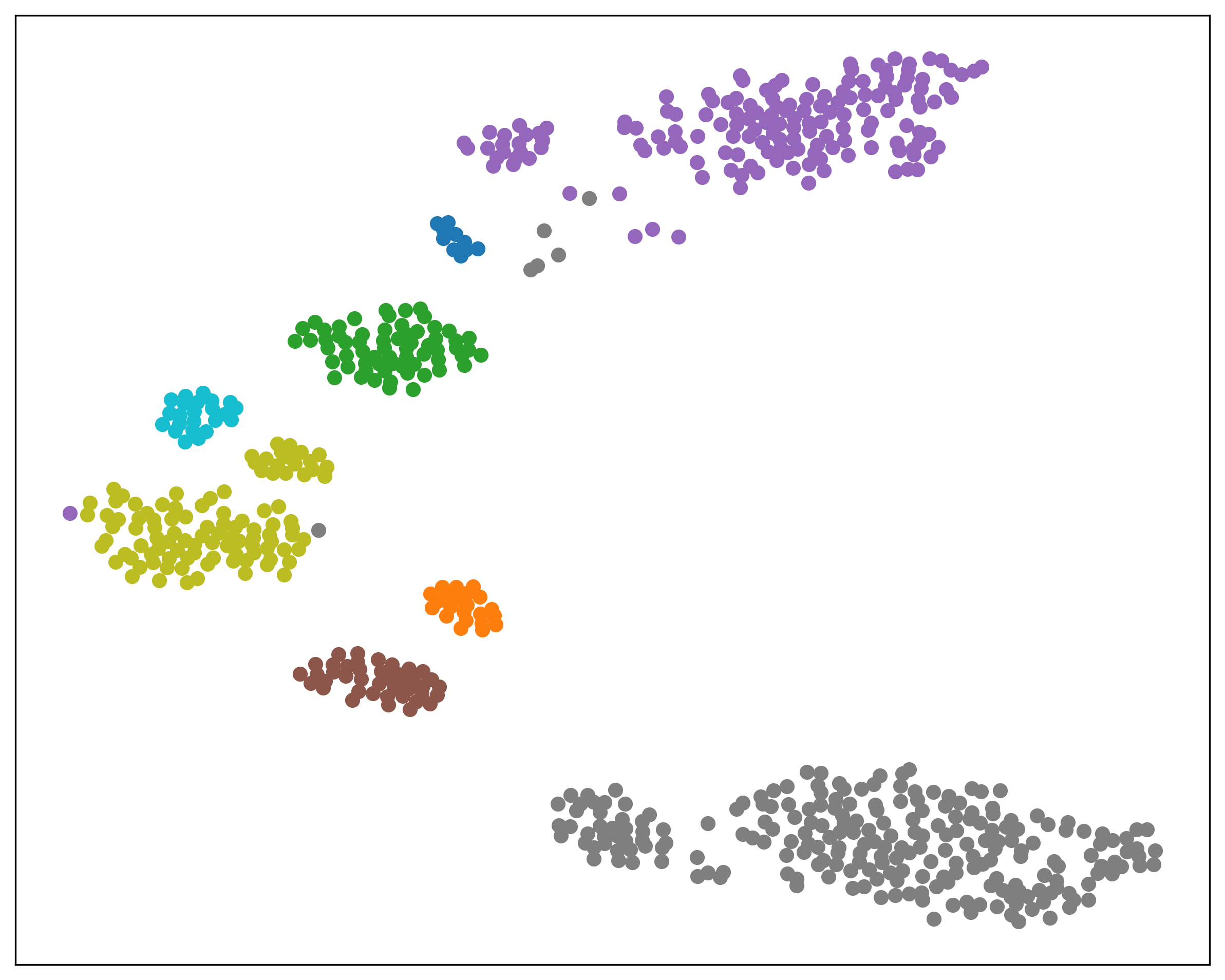}
    \end{subfigure}
    \caption{\textbf{Identity disentanglement visualization.}
    \textbf{Top}: Examples of different people with similar outfits. These ID groups are used for identity feature visualization in bottom row.
    \textbf{Bottom}: Identity feature cluster visualization results, with baseline results on the left and DIFFER results on the right. Different colors represent different person IDs. The baseline features exhibit greater dispersion, whereas the features produced by DIFFER demonstrate tighter clustering within individual identity groups.
    }
    \label{fig:id cluster}
\end{figure}

We visualize the disentangled identity image feature with t-SNE \cite{van2008visualizing} of the same clothing classes from the clothing textual feature cluster in \cref{fig:id cluster}. The LTCC test dataset is used. As shown in the first row in \cref{fig:id cluster}, all the identities are dressed in a similar all-black style. Features from the baseline method are more dispersed without clear boundaries between different identity classes. On the contrary, the features from different ID groups in the proposed method are separated with clear boundaries. This demonstrates the proposed disentangle method successfully differentiates the identity feature from the encoded image feature space without the influence of similar clothing features.

\section{Examples of enhanced performance on baseline failure cases}

\begin{figure*}[t!]
    \centering
    \begin{subfigure}{0.09\textwidth}
        \centering
        \includegraphics[height=3.5cm]{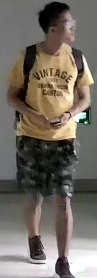}
         \caption*{a}
    \end{subfigure}
    \hspace{0.001\textwidth}
    \begin{subfigure}{0.09\textwidth}
        \centering
        \includegraphics[height=3.5cm]{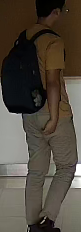}
        \caption*{b}
    \end{subfigure}
    \hspace{0.001\textwidth}
    \begin{subfigure}{0.09\textwidth}
        \centering
        \includegraphics[height=3.5cm]{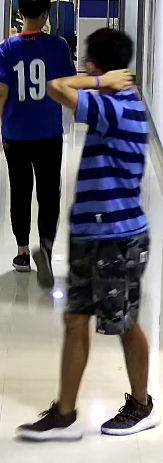}
        \caption*{c}
    \end{subfigure}
    \hspace{0.05\textwidth}
    \begin{subfigure}{0.09\textwidth}
        \centering
        \includegraphics[height=3.5cm]{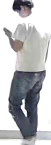}
        \caption*{a}
    \end{subfigure}
    \hspace{0.001\textwidth}
    \begin{subfigure}{0.09\textwidth}
        \centering
        \includegraphics[height=3.5cm]{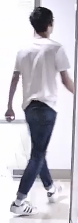}
        \caption*{b}
    \end{subfigure}
    \hspace{0.001\textwidth}
    \begin{subfigure}{0.09\textwidth}
        \centering
        \includegraphics[height=3.5cm]{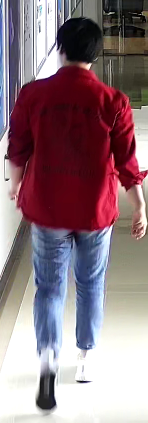}
        \caption*{c}
    \end{subfigure}
    \hspace{0.05\textwidth}
    \begin{subfigure}{0.09\textwidth}
        \centering
        \includegraphics[height=3.5cm]{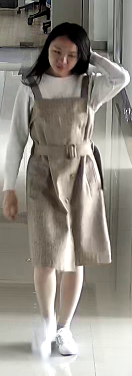}
        \caption*{a}
    \end{subfigure}
    \hspace{0.001\textwidth}
    \begin{subfigure}{0.09\textwidth}
        \centering
        \includegraphics[height=3.5cm]{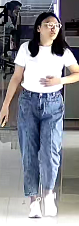}
        \caption*{b}
    \end{subfigure}
    \hspace{0.001\textwidth}
    \begin{subfigure}{0.09\textwidth}
        \centering
        \includegraphics[height=3.5cm]{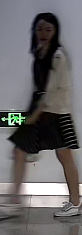}
        \caption*{c}
    \end{subfigure}

     \vspace{0.3cm}

     \begin{subfigure}{0.09\textwidth}
        \centering
        \includegraphics[height=3.5cm]{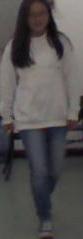}
         \caption*{a}
    \end{subfigure}
    \hspace{0.001\textwidth}
    \begin{subfigure}{0.09\textwidth}
        \centering
        \includegraphics[height=3.5cm]{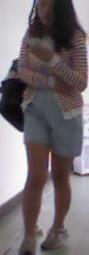}
        \caption*{b}
    \end{subfigure}
    \hspace{0.001\textwidth}
    \begin{subfigure}{0.09\textwidth}
        \centering
        \includegraphics[height=3.5cm]{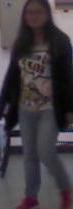}
        \caption*{c}
    \end{subfigure}
    \hspace{0.05\textwidth}
    \begin{subfigure}{0.09\textwidth}
        \centering
        \includegraphics[height=3.5cm]{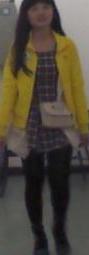}
        \caption*{a}
    \end{subfigure}
    \hspace{0.001\textwidth}
    \begin{subfigure}{0.09\textwidth}
        \centering
        \includegraphics[height=3.5cm]{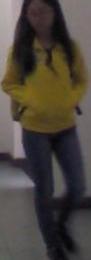}
        \caption*{b}
    \end{subfigure}
    \hspace{0.001\textwidth}
    \begin{subfigure}{0.09\textwidth}
        \centering
        \includegraphics[height=3.5cm]{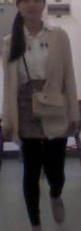}
        \caption*{c}
    \end{subfigure}
    \hspace{0.05\textwidth}
    \begin{subfigure}{0.09\textwidth}
        \centering
        \includegraphics[height=3.5cm]{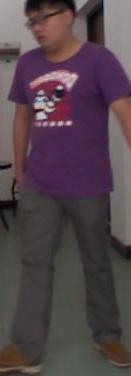}
        \caption*{a}
    \end{subfigure}
    \hspace{0.001\textwidth}
    \begin{subfigure}{0.09\textwidth}
        \centering
        \includegraphics[height=3.5cm]{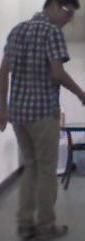}
        \caption*{b}
    \end{subfigure}
    \hspace{0.001\textwidth}
    \begin{subfigure}{0.09\textwidth}
        \centering
        \includegraphics[height=3.5cm]{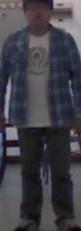}
        \caption*{c}
    \end{subfigure}

     \vspace{0.3cm}
     
    \begin{subfigure}{0.09\textwidth}
        \centering
        \includegraphics[height=3.5cm]{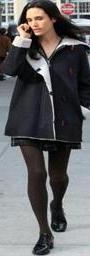}
         \caption*{a}
    \end{subfigure}
    \hspace{0.001\textwidth}
    \begin{subfigure}{0.09\textwidth}
        \centering
        \includegraphics[height=3.5cm]{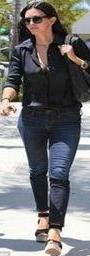}
        \caption*{b}
    \end{subfigure}
    \hspace{0.001\textwidth}
    \begin{subfigure}{0.09\textwidth}
        \centering
        \includegraphics[height=3.5cm]{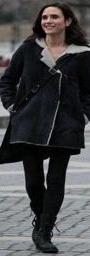}
        \caption*{c}
    \end{subfigure}
    \hspace{0.05\textwidth}
    \begin{subfigure}{0.09\textwidth}
        \centering
        \includegraphics[height=3.5cm]{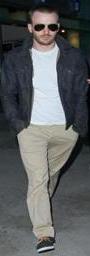}
        \caption*{a}
    \end{subfigure}
    \hspace{0.001\textwidth}
    \begin{subfigure}{0.09\textwidth}
        \centering
        \includegraphics[height=3.5cm]{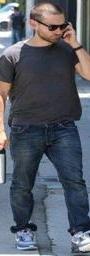}
        \caption*{b}
    \end{subfigure}
    \hspace{0.001\textwidth}
    \begin{subfigure}{0.09\textwidth}
        \centering
        \includegraphics[height=3.5cm]{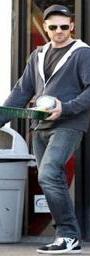}
        \caption*{c}
    \end{subfigure}
    \hspace{0.05\textwidth}
    \begin{subfigure}{0.09\textwidth}
        \centering
        \includegraphics[height=3.5cm]{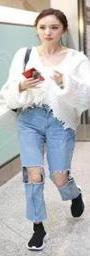}
        \caption*{a}
    \end{subfigure}
    \hspace{0.001\textwidth}
    \begin{subfigure}{0.09\textwidth}
        \centering
        \includegraphics[height=3.5cm]{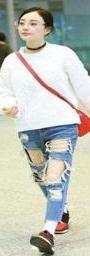}
        \caption*{b}
    \end{subfigure}
    \hspace{0.001\textwidth}
    \begin{subfigure}{0.09\textwidth}
        \centering
        \includegraphics[height=3.5cm]{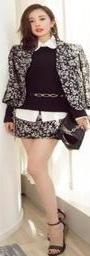}
        \caption*{c}
    \end{subfigure}
    
    \caption{\textbf{Example of improvement of DIFFER on different datasets}. From the top to bottom rows are examples from LTCC, PRCC, and Celeb-reID-light respectively. (a: query image; b: baseline method top1 matched error image; c: DIFFER top1 matched correct image)}
    \label{fig:errorMatch}
\end{figure*}

This section illustrates the effectiveness of DIFFER compared to the baseline model through examples from three benchmark datasets: LTCC, PRCC, and Celeb-reID-light, as shown in \cref{fig:errorMatch}. Each row corresponds to examples from one dataset, where (a) represents the query image, (b) shows the top-1 incorrect match by the baseline model, and (c) displays the top-1 correct match by DIFFER.

As seen in the examples, DIFFER demonstrates its ability to address significant challenges that often hinder the baseline model. These challenges include changes in clothing, where individuals are dressed in entirely different outfits; similar poses, where non-biometric factors such as body orientation or posture confuse the baseline model; and overlapping characteristics like hairstyles or environmental settings that lead to false matches. For instance, in the LTCC dataset examples, DIFFER correctly matches individuals despite substantial changes in their attire, where the baseline confuses individuals with similar clothing patterns. Similarly, in the PRCC dataset, DIFFER handles challenging cases where hairstyle similarity between different individuals misleads the baseline. In the Celeb-reID-light examples, DIFFER effectively overcomes confounding factors such as varying poses and subtle similarities in appearance that the baseline fails to disentangle.

These results highlight DIFFER's robustness in disentangling biometric information from non-biometric factors and its ability to leverage semantic descriptions effectively. By successfully overcoming the limitations of the baseline model, DIFFER significantly improves identification accuracy across diverse and challenging scenarios. This demonstrates the practical applicability of the proposed method in real-world settings where non-biometric interference is prevalent.

\end{document}